# Prediction of Zoonosis Incidence in Human using Seasonal Auto Regressive Integrated Moving Average (SARIMA)

Adhistya Erna Permanasari
Computer and Information Science Dept.
Universiti Teknonologi PETRONAS
Bandar Seri Iskandar, 31750 Tronoh,
Perak, Malaysia
astya_00@yahoo.com

Dayang Rohaya Awang Rambli
Computer and Information Science Dept.
Universiti Teknonologi PETRONAS
Bandar Seri Iskandar, 31750 Tronoh,
Perak, Malaysia
roharam@petronas.com.my

P. Dhanapal Durai Dominic
Computer and Information Science Dept.
Universiti Teknonologi PETRONAS
Bandar Seri Iskandar, 31750 Tronoh,
Perak, Malaysia
Dhanapal_d@petronas.com.my

*Abstract*— Zoonosis refers to the transmission of infectious diseases from animal to human. The increasing number of zoonosis incidence makes the great losses to lives, including humans and animals, and also the impact in social economic. It motivates development of a system that can predict the future number of zoonosis occurrences in human. This paper analyses and presents the use of Seasonal Autoregressive Integrated Moving Average (SARIMA) method for developing a forecasting model that able to support and provide prediction number of zoonosis human incidence. The dataset for model development was collected on a time series data of human Salmonellosis occurrences in United States which comprises of fourteen years of monthly data obtained from a study published by Centers for Disease Control and Prevention (CDC). Several trial models of SARIMA were compared to obtain the most appropriate model. Then, diagnostic tests were used to determine model validity. The result showed that the SARIMA(9,0,14)(12,1,24)$_{12}$ is the fittest model. While in the measure of accuracy, the selected model achieved 0.062 of Theil's U value. It implied that the model was highly accurate and a close fit. It was also indicated the capability of final model to closely represent and made prediction based on the tuberculosis historical dataset.

I. INTRODUCTION

Zoonosis refers to any infectious disease that is transmitted from animals humans [1, 2]. It was estimated that around 75% of emerging disease infections to humans come from animal origin [3-5]. The zoonosis epidemics arise and exhibit the potential threat for public health and economic impact. Large numbers of people have been killed by zoonotic disease in different countries.

The WHO statistic [6] reported some zoonosis outbreaks including Dengue/dengue haemorrhagic fever in Brazil (647 cases, with 48 deaths); Avian Influenza outbreaks in 15 countries (438 cases, 262 deaths) ; Rift Valley Fever in Sudan (698 cases, including 222 deaths); Ebola in Uganda (75 patients), Ebola in Philippines (6 positive cases from 141 suspect) and Ebola in Congo (32 cases, 15 deaths); and the latest was Swine Flu (H1N1) in many countries (over 209438 cases, at least 2185 deaths). Some of these zoonoses recently have major outbreaks worldwide which resulted in many losses of lives both to humans and animals.

Zoonosis evolution from the original form could cause the newly emerging zoonotic disease [4]. Indeed this is evidenced in a report presented by WHO [4] associating microbiological factors with the agent, the animal hosts/reservoirs and the human victims which could result in a new variant of a pathogen that is capable of jumping the species barrier. For example, Influenza A virus mechanism have jumped from wild waterfowl species into domestic farm, farm animal, and humans. The other recent example is the swine flu that the outbreaks in human originally come from a new influenza virus in a swine. The outbreak of disease in people caused by a new influenza virus of swine origin continues to grow in the United States and internationally

Worldwide frequency of zoonosis outbreak in the past 30 years [3] and the risk factor of the newly emerging diseases forced many governments to apply stringent measures to prevent zoonosis outbreak, for example by destroying the last livestock in the infected area. These mean great losses to farmer. The significant impact to human life, however, still remains the biggest issue in zoonosis. Therefore, it highlights the need for a modeling approach that can give decision makers an early estimate of future number zoonosis incidence, based on the historical time series data. The use of computer software couple with a statistical modeling can be used to forecast the number of zoonosis incidence.

Time series analysis regarding forecasting model is widely used in various fields. In fact, there are few of studies regarding zoonosis forecasting comparing to other areas, such as energy demand prediction, economic field, traffic prediction, and in the health support. Indeed, prediction the risk of zoonosis impact in human need to be focused, due to the need to obtain the result to take the further decision.

Many researchers have developed different forecasting methods to predict zoonosis human incidence.

Multivariate Markov chain model was selected to project the number of tuberculosis (TB) incidence in the United States from 1980 to 2010 [7]. This work pointed out the study of TB incidence based on demographic groups. The uncertainty in model parameters was handled by fuzzy number. The model determined that the decline rate in the number of cases among Hispanics would be slower than among white non-Hispanics and black non-Hispanics.

The number of human incidence of Schistosoma haematobium at Niono, Mali was forecasted online by using exponential smoothing method [8]. The method was used as a core of a proposed state-space framework. The data was collected from 1996-2004 from 17 community health center in that area. The framework was able to assist managing and assessing S. haematobium transmission and intervention impact,

Three different approaches were applied to forecast the SARS epidemic in China. The existing time series was processed by AR(1), ARIMA(0,1,0), and ARMA(1,1). The result could be used to support the disease reports [9]. The result of this study could be used to monitor the dynamic of SARS in China based on the daily data.

A Bayesian dynamic model also could be used to to monitor the influenza surveillance as one factor of SARS epidemic [10]. This model was developed to link pediatric and adult syndromic data to the traditional measures of influenza morbidity and mortality. The findings showed the importance of modeling influenza surveillance data, and recommend dynamic Bayesian Network.

Monthly data of Cutaneous Leishmaniasis (CL) incidence in Costa Rica from 1991 to 2001 was analyzed by using seasonal autoregressive models. This work was studying the relationship between the interannual cycles of the diseases with the climate variables using frequency and time-frequency techniques of time series analysis [11]. This model supported the dynamic link between the disease and climate.

Additive decomposition method was used to predict Salmonellosis incidence in US [12]. Fourteen years historical data from 1993 to 2006 was collected to compute the forecast values until 12 months-ahead.

Different forecasting approaches were applied into zoonosis time series. However, the increasing number of zoonosis occurrences in human made the need to take a further study of zoonosis forecasting in different zoonotic disease [13]. Due to that issue, selections of the fitted model were necessary to obtain the optimal result.

This paper analyses the empirical results for evaluating and predicting the number of zoonosis incidence by using Autoregressive Integrated Moving Average (ARIMA). This model is selected because of the capability to correct the local trend in data, where the pattern in the previous period can be used to forecast the future. Thus this model also supports in modeling one perspective as a function of time (in this case, the number of human case) [14]. Due to the seasonal trend of time series used, the Seasonal ARIMA (SARIMA) is selected for the model development.

The remainder of the paper is structured as follows. Section II presents preparation of the time series. Section III describes basic theory of ARIMA and SARIMA model. Section IV introduces Bayesian Information Criterion (BIC) and Akaike Information Criterion (AIC). Section V reports model development. Finally, Section VI present conclusion and directions for future work.

II. DATASET FOR MODEL DEVELOPMENT

This section describes the dataset (time series) that was used for model development. Salmonellosis disease was selected because these incidences can found in any country. A study collected time series data of human Salmonellosis occurrences in United State for the 168 month period from January 1993 to December 2006. The data was obtained from the summary of notifiable diseases in United States from the Morbidity and Mortality Weekly Report (MMWR) that published by Centers for Disease Control and Prevention (CDC). The seasonal variation of the original data is presented in Fig. 1. Then, trend in every month is plotted by using seasonal stacked line in Fig.2.

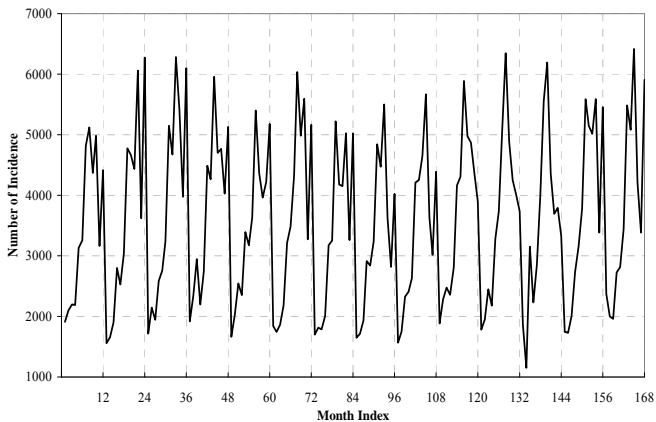

Figure 1. Monthly number of US Salmonellosis incidence (1993-2006)

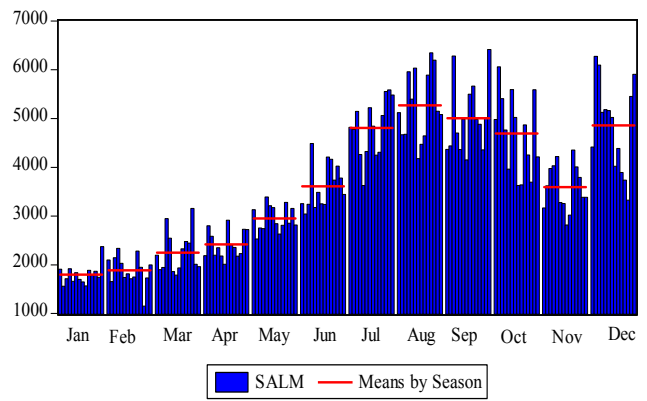

Figure 2. Seasonal stacked line of US Salmonellosis (1993-2006)

Fig. 2 shows seasonal stacked line for human Salmonellosis incidence in US from 1993 until 2006. The plot shows a peak season of incidence in August while the minimum number of incidence occurrences in January. Since time series plot of the historical data exhibited the seasonal variations which present similar trend every year, then SARIMA was chosen as the appropriate approach to develop a model prediction.

## III. ARIMA AND SEASONAL ARIMA MODEL

This section introduces the basic theory of Autoregressive Integrated Moving Average (ARIMA). The general class of ARIMA (*p,d,q*) processes shown in (1) as

$$y_t = \delta + \phi_1 y_{t-1} + \phi_2 y_{t-2} + \ldots + \phi_p y_{t-p} \\ + a_t - \theta_1 a_{t-1} - \theta_2 a_{t-2} - \ldots - \theta_q a_{t-q} \quad (1)$$

where *d* is the level of differencing, *p* is the autoregressive order, and *q* is the moving average order [15]. The constant is notated by *δ*, while *φ* is an autoregressive operator and *θ* is a moving average operator.

Seasonal ARIMA (SARIMA) is used when the time series exhibits a seasonal variation. A seasonal autoregressive notation (*P*) and a seasonal moving average notation (*Q*) will form the multiplicative process of SARIMA as (*p,d,q*)(*P,D,Q*)$_s$. The subscripted letter 's' shows the length of seasonal period. For example, in a hourly data time series *s* = 7, in a quarterly data *s* = 4, and in a monthly data *s* = 12.

In order to formalize the model, the *backshift* operator (*B*) is used. The time series observation backward in time by *k* period is symbolized by $B^k$, such that $B^k y_t = y_{t-k}$

Formerly, the backshift operator is used to present a general stationarity transformation, where the time series is stationer if the statistical properties (mean and variance) are constant through time. The general stationarity transformation is presented below:

$$z_t = \nabla_s^D \nabla^d y_t = (1 - B^s)^D (1 - B)^d y_t \quad (2)$$

where *z* is the time series differencing, *d* is the degree of nonseasonal differencing used and *D* is the degree of seasonal differencing used.
Then, the general SARIMA (*p,P,q,Q*) model is

$$\phi_p(B)\phi_P(B^s) z_t = \delta + \theta_q(B)\theta_Q(B^s) a_t \quad (3)$$

Where:

- $\phi_p(B) = (1 - \phi_1 B - \phi_2 B^2 - \ldots - \phi_p B^p)$
  is the nonseasonal autoregressive operator of order *p*
- $\phi_P(B^L) = (1 - \phi_{1,L} B^L - \phi_{2,L} B^{2L} - \ldots - \phi_{P,L} B^{PL})$ is the seasonal autoregressive operator of order *P*
- $\theta_q(B) = (1 - \theta_1 B - \theta_2 B^2 - \ldots - \theta_q B^q)$
  is the nonseasonal moving average operator of order *q*
- $\theta_Q(B^L) = (1 - \theta_{1,L} B^L - \theta_{2,L} B^{2L} - \ldots - \theta_{Q,L} B^{QL})$
  is the seasonal moving average operator of order *Q*
- $\delta = \mu \phi_p(B) \phi_P(B^L)$
  is a constant term where *μ* is the mean of stationary time series
- $\phi, \theta, \delta$ are unknown parameter that can be calculated from the sample data.
- $a_t, a_{t-1}, \ldots$ are random shocks that are assumed to be independent of each other

George Box and Gwilym Jenkins studied the simplified step to obtain the comprehensive information of understanding ARIMA model and using the univariate ARIMA model [15],[16]. The Box-Jenkins (BJ) methodology consists of four iterative steps:

*1) Step 1: Identification*
This step focus on selection of the order of regular differencing (*d*), seasonal differencing (*D*), the non-seasonal order of Autoregressive (*p*), the seasonal order of Autoregressive (*P*), the non-seasonal order of Moving Average (*q*) and the non-seasonal order of Autoregressive (*Q*). The number of order can be identified by observing the sample autocorrelations (SAC) and sample partial autocorrelations (SPAC).

*2) Step 2: Estimation*
The historical data is used to estimate the parameters of the tentatively model in Step 1.

*3) Step 3: Diagnostic checking*
Various diagnostic tests are used to check the adequacy of the tentatively model.

*4) Step 4: Forecasting*
The final model in step 3 then is used to forecast the forecast values.

This approach is widely used to examining the SARIMA model because of the capability to capture the appropriate trend by examining historical pattern. The BJ methodology has several advantages, involving extract a great deal of information from the time series using a minimum number of parameters and the capability in handling stationery and non-stationary time series in non-seasonal and seasonal elements [17],[18].

## IV. BAYESIAN INFORMATION CRITERION (BIC) AND AKAIKE INFORMATION CRITERION (AIC)

Selection of ARIMA model was based on the Bayesian Information Criterion (BIC) and Akaike Information Criterion (AIC) values. These models are using Maximum Likelihood principle to choose highest possible dimension. The determinant of the residual covariance is computed as:

$$|\hat{\Omega}| = \det\left(\frac{1}{T-p} \sum_t \hat{\varepsilon}_t \hat{\varepsilon}_t' / T\right) \quad (4)$$

The log likelihood value is assuming computed by a multivariate normal (Gaussian) distribution as:

$$l = -\frac{T}{2}\{k(1+\log 2\pi) + \log|\hat{\Omega}|\} \quad (5)$$

Then the AIC and BIC are formulated as [19]:

$$AIC = -2(l/T) + 2(n/T) \quad (6)$$

$$BIC = -2(l/T) + n\log(T)/T \quad (7)$$

where $l$ is the value of the log of the likelihood function with the $k$ parameters estimated using $T$ observations and $n = k(d + pk)$. The various information criteria are all based on –2 times the average log likelihood function, adjusted by a penalty function.

V. MODEL DEVELOPMENT

This following section discusses the result of BJ iterative steps to forecast an available dataset.

A. Identification

Starting with BJ methodology introduced in section 3, the first step in the model development is to identify the dataset. In this step, sample autocorrelations (SAC) and sample partial autocorrelations (SPAC) of the historical data were plotted to observe the pattern.

Three periodical data was selected to illustrate the plot. The result is shown in Fig. 3. Based on Fig. 3, it could be observed that the correlogram of time series is likely to have seasonal cycles especially in SAC which implied level non-stationary. Therefore, the regular differencing and seasonal differencing was applied to the original time series as presented in Fig. 4 and Fig. 5.

An Augmented Dickey-Fuller (ADF) test was performed to determine whether a data differencing is needed [19]. The null hypothesis of the Augmented Dickey-Fuller t-test is:

- $H_0 : \theta = 0$ then the data needs to be differenced to make it stationary, versus the alternative hypothesis of

- $H_1 : \theta < 0$ then the data is stationary and doesn't need to be differenced

The result was compared with the 1%, 5%, and 10% critical values to indicate non-rejection of the null hypothesis. The ADF test statistic value had a t-Statistic value of -1.779 and the one-sided $p$-value is 0.389. The critical values reported at 1%, 5%, and 10% were -3.476, -2.882, -2.578. It showed that $t_\alpha$ value was greater than the critical values that provide evidence not to reject the null hypothesis of a unit root then the time series need to be differencing.

The regular differencing and seasonal differencing was applied to the original time series. The ADF test also applied for both of them. The result showed the critical values of the regular differencing were -14.171 and for the seasonal differencing were -12.514. The one-sided $p$-value for both differencing was 0.000. While the probability value of 0.000 provided evidence to reject the null hypotheses. It indicated the stationarity of the time series.

Figure 3. SAC and SPAC correlogram of original data

Figure 4. SAC and SPAC correlogram of regular differencing

Figure 5. SAC and SPAC correlogram of seasonal differencing

Selecting of whether to use regular or seasonal differencing was based on the correlogram. In order to develop ARIMA, time series should be stationary. Based on the correlogram shown in Figure 3 and Figure 4 was observed that more spikes was found in the regular differencing than the seasonal differencing. Then, the seasonal differencing was chosen for model development.

*B. Parameter Estimation*

Different ARIMA models were applied to find the best fitting model. The most appropriate model was selected by using BIC and AIC values. The best model was determined from the minimum BIC and AIC. Table I presents the results of estimating the various ARIMA processes for the seasonal differencing of Salmonellosis human incidence using the EViews 5.1 econometric software package.

TABLE I. ESTIMATION OF SELECTED SARIMA MODEL

| No | Model Variable | BIC | AIC | Adj. $R^2$ |
|---|---|---|---|---|
| 1 | C, AR(9), SAR(12), SAR(22), SAR(24), MA(9), SMA(12), SMA(24) | 15.614 | 15.431 | 0.345 |
| 2 | AR(9), SAR(12), SAR(22), SAR(24), MA(9), SMA(12), SMA(24) | 15.587 | 15.427 | 0.342 |
| 3 | AR(9), SAR(12), SAR(24), MA(9),SMA(12), SMA(22), SMA(24) | 15.583 | 15.423 | 0.345 |
| 4 | AR(3), AR(9), SAR(12), MA(3), SMA(24) | 15.394 | 15.286 | 0.449 |
| 5 | AR(3), AR(9), SAR(12), MA(14), SMA(24) | 15.351 | 15.243 | 0.472 |
| 6 | AR(3), AR(9), SAR(12), MA(24) | 15.368 | 15.282 | 0.447 |
| 7 | AR(9), SAR(12), MA(3), SMA(24) | 15.359 | 15.273 | 0.452 |
| 8 | AR(9), SAR(12), MA(14), SMA(24) | 15.331 | 15.245 | 0.468 |
| 9 | AR(9), SAR(12), MA(3), MA(14), SMA(24) | 15.352 | 15.245 | 0.471 |

The AIC and BIC are commonly used in model selection, whereby the smaller value is preferred. From Table 1, model 8 had a relatively small value of BIC and AIC. It also achieved large adjusted $R^2$.

The model AR(9), SAR(12), MA(14), SMA(24) also could be written as SARIMA(9,0,14)(12,1,24)$_{12}$.

To produce the model, the separated non-seasonal and seasonal model was computed first. It was followed by combining these models to describe the final model.

- Step 1: Model for nonseasonal level
  AR (9) : $z_t = \delta + \phi_9 z_{t-9} + a_t$ (8)
  MA(14) : $z_t = \delta + a_t - \theta_{14} a_{t-14}$ (9)

- Step 2: Model for seasonal level
  AR (12) : $z_t = \delta + \phi_{1,12} z_{t-12} + a_t$ (10)
  MA (24) : $z_t = \delta + a_t - \theta_{2,12} a_{t-24}$ (11)

- Step 3: Combining (8) – (11) has arrived to (12).

$$z_t = \delta + \phi_9 z_{t-9} - \theta_{14} a_{t-14} + \phi_{1,12} z_{t-12} - \theta_{2,12} a_{t-24} + a_t \quad (12)$$

Hence, $\delta$ value was 0.44, less than |2| and statistically not different from zero, then $\delta$ can be excluded from the model. Where an autoregressive and a moving average model presented in the nonseasonal or seasonal level, then multiplicative terms was used as the following:

- $\phi_9 z_{t-9}$ and $\phi_{1,12} z_{t-12}$ was used to form the multiplicative term $\phi_9 \phi_{1,12} z_{t-21}$

- $-\theta_{14} a_{t-14}$ and $-\theta_{2,12} a_{t-24}$ was used to form the multiplicative term $\theta_{14} \theta_{2,12} a_{t-38}$

The model was derived using multiplicative form as follows:

$$\begin{aligned} z_t &= \phi_9 z_{t-9} - \theta_{14} a_{t-14} + \phi_{1,12} z_{t-12} - \theta_{2,12} a_{t-24} \\ &\quad + \phi_9 \phi_{1,12} z_{t-21} + \theta_{14} \theta_{2,12} a_{t-38} + a_t \end{aligned} \quad (13)$$

$$\begin{aligned} z_t - \phi_9 z_{t-9} - \phi_{1,12} z_{t-12} - \phi_9 \phi_{1,12} z_{t-21} \\ = \theta_{14} a_{t-14} + \theta_{2,12} a_{t-24} - \theta_{14} \theta_{2,12} a_{t-38} + a_t \end{aligned}$$

The backshift operator (*B*) was applied in (13) yield:

$$\begin{aligned} z_t - \phi_9 B^9 z_t - \phi_{1,12} B^{12} z_t - \phi_9 \phi_{1,12} B^{21} z_t \\ = \theta_{14} B^{14} a_t + \theta_{2,12} B^{24} a_t - \theta_{14} \theta_{2,12} B^{38} a_t + a_t \end{aligned} \quad (14)$$

$$\begin{aligned} (1 - \phi_9 B^9 - \phi_{1,12} B^{12} - \phi_9 \phi_{1,12} B^{21}) z_t \\ = (1 + \theta_{14} B^{14} + \theta_{2,12} B^{24} - \theta_{14} \theta_{2,12} B^{38}) a_t \end{aligned}$$

From the computation the parameter result were AR(9) = 0.154, SAR(12) = -0.513, MA(14) = 0.255, SMA(24) = -0.860. The estimated parameters were included into (14) to form the final model that expressed as follows:

$$\begin{aligned} (1 - 0.154 B^9 + 0.513 B^{12} + 0.078 B^{21}) z_t \\ = (1 + 0.255 B^{14} - 0.860 B^{24} - 0.219 B^{38}) a_t \end{aligned} \quad (15)$$

Since the seasonal differencing was chosen, then (2) was notated with d = 0, D = 1 and s = 12 to define $z_t$ as:

$$\begin{aligned} z_t &= \nabla_s^D \nabla^d y_t = (1 - B^s)^1 (1 - B)^0 y_t \\ &= (1 - B^s) y_t = y_t - B^s y_t = y_t - y_{t-s} \\ &= y_t - y_{t-12} \end{aligned} \quad (16)$$

The SARIMA final model was used to compute the forecast values for the three years-ahead.

*C. Diagnostic Checking*

Diagnostic check was made into the selected model. Correlogram and the residual plots are presented in Fig. 6. LM (Lagrange multiplier) Test was applied for the first lag period (lag 1 – lag 12) where the result can be seen in Table 2. There is no serial correlation in the residual because the SAC and SPAC at all lag nearly zero and within the 95% confidence interval.

Figure 6. SAC and SPAC correlogram of model residual

TABLE II. LM TEST RESULT FOR RESIDUAL

| Variable | Std. Error | t-Statistic | Prob. |
|---|---|---|---|
| AR(9) | 0.224 | 0.510 | 0.611 |
| SAR(12) | 0.124 | -0.804 | 0.423 |
| MA(14) | 0.084 | -0.049 | 0.961 |
| SMA(24) | 0.029 | -0.223 | 0.824 |
| RESID(-1) | 0.093 | -0.225 | 0.822 |
| RESID(-2) | 0.092 | 1.252 | 0.213 |
| RESID(-3) | 0.093 | 1.448 | 0.150 |
| RESID(-4) | 0.095 | -1.345 | 0.181 |
| RESID(-5) | 0.096 | 0.730 | 0.467 |
| RESID(-6) | 0.096 | -1.141 | 0.256 |
| RESID(-7) | 0.098 | -0.547 | 0.586 |
| RESID(-8) | 0.096 | 0.036 | 0.971 |
| RESID(-9) | 0.224 | -0.250 | 0.803 |
| RESID(-10) | 0.096 | -0.647 | 0.274 |
| RESID(-11) | 0.094 | 0.178 | 0.721 |
| RESID(-12) | 0.156 | 0.943 | 0.432 |

Refer to the result from Table II, it proved that there were no correlation up to order 12 because the t-Statistic for SAC and SPAC less than |2|. Then it was concluded that the selected model was fit.

### D. Forecasting

SARIMA$(9,0,14)(12,1,24)_{12}$ was selected as the most appropriate model from various traces. It was used to forecast the predicted incidence from 2007 through 2009 ($t_{169} - t_{204}$) that present in the Table III. A completed time series plots in Fig. 7 which consists of three components: actual values, fitted values, and residual values.

TABLE III. THE FORECASTING RESULT.

| Month | Prediction | | |
|---|---|---|---|
| | 2007 | 2008 | 2009 |
| January | 1678.571 | 1678.544 | 1678.558 |
| February | 1965.707 | 1965.624 | 1965.666 |
| March | 1969.415 | 1969.383 | 1969.399 |
| April | 2692.214 | 2692.268 | 2692.240 |
| May | 2657.628 | 2657.578 | 2657.604 |
| June | 3364.616 | 3364.677 | 3364.646 |
| July | 5020.626 | 5020.563 | 5020.595 |
| August | 4675.720 | 4675.721 | 4675.720 |
| September | 5655.739 | 5655.426 | 5655.587 |
| October | 5691.898 | 5691.992 | 5691.944 |
| November | 3489.930 | 3489.962 | 3489.946 |
| December | 5639.062 | 5639.223 | 5639.140 |

### E. Error Measures

After empirical examination, forecast accuracy was initially calculated using different accuracy measures: RMSE, MAD, MAPE, and Theil's U-statistic. The result showed that RMSE was RMSE was 479.99, MAD was 367.23, MAPE was 10.36, and Theil's U was 0.062. On hindsight RMSE, MAD, and MAPE are stand-alone accuracy measures that have inherent disadvantages that could lead to certain losses of functions [20]. The use of relative measures however could resolve this limitation. Instead of naïve forecasting, relative measures evaluate the performance of a forecast. Moreover, it can eliminate the bias from trends, seasonal components, and outliers. As such, forecast accuracy for the selected model is based on relative accuracy measure, Theil's U statistic. Based on the Theil's U-statistics of value 0.062, the model is highly accurate and present a close fit. Thus, the empirical result indicated that the model was able to accurately represent the Salmonellosis historical dataset.

### VI. CONCLUSION

In this paper, the use of forecasting method was applied to predict the number of Salmonellosis human incidence in US based on the monthly data. The adjusted model prediction was developed by using SARIMA model based on the historical data. Different SARIMA model was tested to select an appropriate ARIMA model. Various diagnostic check were used to determine model validity, including BIC and AIC. The result indicate that SARIMA$(9,0,14)(12,1,24)_{12}$ was the fittest model among them.

The empirical result indicated that the model was able to represent the historical data with Theil's U with the value 0,062. In addition, SARIMA model can be obtained by using four iteratively Box-Jenkins steps and provide the prediction of the number of human incidence in other zoonosis to help the stakeholder make further decision. A further work is still needed to evaluate and apply other forecasting methods into the zoonosis time series in order to obtain better accuracy of forecast value.

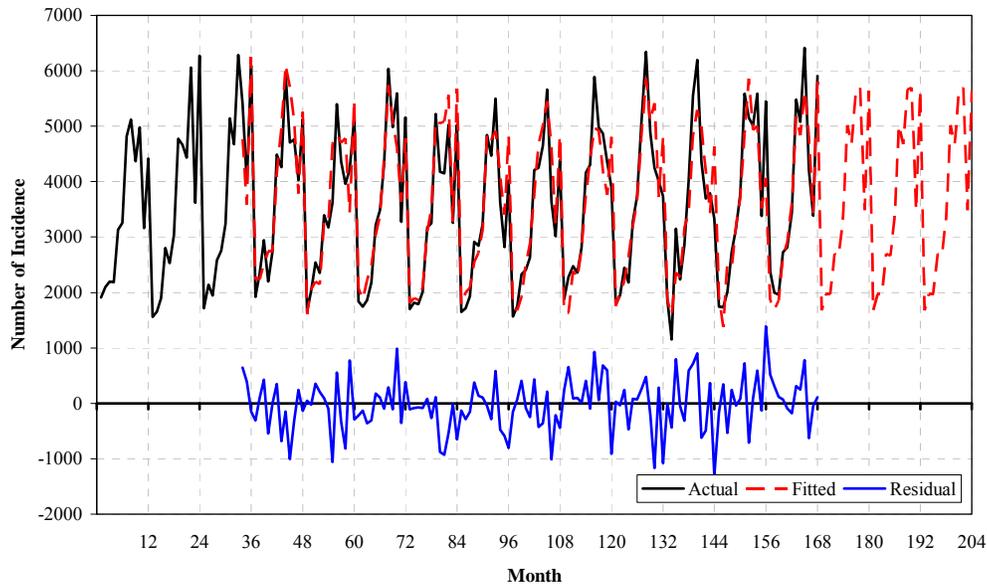

Figure 7. Time series forecasting plot

REFERENCES

[1] CDC, "Compendium of Measures To Prevent Disease Associated with Animals in Public Settings," National Association of State Public Health Veterinarians, Inc. (NASPHV) MMWR 2005;54 (No. RR-4), 2005.

[2] WHO. (2007). Zoonoses and veterinary public health (VPH) [Online]. Available: http://www.who.int

[3] B. A. Wilcox and D. J. Gubler, "Disease Ecology and the Global Emergence of Zoonotic Pathogens," *Environmental Health and Preventive Medicine*, vol. 10, pp. 263–272, 2005.

[4] WHO, "Report of the WHO/FAO/OIE joint consultation on emerging zoonotic diseases," Geneva, Switzerland, 3–5 May 2004.

[5] J. Slingenbergh, M. Gilbert, K. de Balogh, and W. Wint, "Ecological sources of zoonotic diseases," *Rev. sci. tech. Off. int. Epiz.*, vol. 23, pp. 467-484, 2004.

[6] WHO. (2009). Disease Outbreak News [Online]. Available: http://www.who.int/csr/don/en/

[7] S. M. Debanne, R. A. Bielefeld, G. M. Cauthen, T. M. Daniel, and D. Y. Rowland, "Multivariate Markovian Modeling of Tuberculosis: Forecast for the United States," *Emerging Infectious Diseases*, vol. 6, 2000.

[8] D. C. Medina, S. E. Findley, and S. Doumbia, "State–Space Forecasting of Schistosoma haematobium Time-Series in Niono, Mali," *PLoS Neglected Tropical Diseases*, vol. 2, 2008.

[9] D. Lai, "Monitoring the SARS Epidemic in China: A Time Series Analysis," *Journal of Data Science*, vol. 3, pp. 279-293, 2005.

[10] P. Sebastiani, K. D. Mandl, P. Szolovits, I. S. Kohane, and M. F. Ramoni, "A Bayesian Dynamic Model for Influenza Surveillance," *Statistics in Medicine*, vol. 25, pp. 1803-1825, 2006.

[11] L. F. Chaves and M. Pascual, "Climate Cycles and Forecasts of Cutaneous Leishmaniasis, a Nonstationary Vector-Borne Disease," *PLoS Medicine*, vol. 3 (8), pp. 1320-1328, 2006.

[12] A. E. Permanasari, D. R. Awang Rambli, and D. D. Dominic, "Forecasting of Zoonosis Incidence in Human Using Decomposition Method of Seasonal Time Series," in *Proc NPC 2009*, 2009, pp. 1-7.

[13] A. E. Permanasari, D. R. Awang Rambli, D. D. Dominic, and V. Paruvachi Ammasa, "Construction of Zoonosis Domain Relationship as a Preliminary Stage for Developing a Zoonosis Emerging System," in *Proc ITSim 2008*, Kuala Lumpur, 2008, pp. 527-534.

[14] E. S. Shtatland, K. Kleinman, and E. M. Cain, "Biosurveillance and outbreak detection using the ARIMA and logistic procedures."

[15] B. L. Bowerman and R. T. O'Connell, *Forecasting and Time Series An Applied Approach*, 3rd ed: Duxbury Thomson Learning, 1993.

[16] S. Makridakis and S. C. Wheelwright, *Forecasting Methods and Applications*: John Wiley & Sons. Inc, 1978.

[17] J. G. Caldwell. (2006). The Box-Jenkins Forecasting Technique [Online]. Available: http://www.foundationwebsite.org/

[18] C. Chia-Lin, S. Songsak, and W. Aree, "Modelling and forecasting tourism from East Asia to Thailand under temporal and spatial aggregation," *Math. Comput. Simul.*, vol. 79, pp. 1730-1744, 2009.

[19] L. Quantitative Micro Software. (2005). EViews 5.1 User's Guide [Online]. Available: www.eviews.com

[20] Z. Chen and Y. Yang. (2004). Assessing Forecast Accuracy Measures [Online]. Available: http://www.stat.iastate.edu/preprint/articles/2004-10.pdf

AUTHORS PROFILE

**Adhistya Erna Permanasari** is a lecturer at the Electrical Engineering Department, Gadjah Mada University, Yogyakarta 55281, Indonesia. She is currently a PhD student at the Department of Computer and Information Science Department, Universiti Teknologi PETRONAS, Malaysia. She received her BS in Electrical Engineering in 2002 and M. Tech (Electrical Engineering) in 2006 from the Department of Electrical Engineering, Gadjah Mada University, Indonesia. Her research interest includes database system, decision support system, and artificial intelligence.

**Dr Dayang Rohaya Awang Rambli** is currently a senior lecturer at the Computer and Information Science Department, Universiti Teknologi PETRONAS, Malaysia. She received her BS from University of Nebraska, M.Sc from Western Michigan University, USA, and Ph.D from Loughborough University, UK. Her primary areas of research interest involve Virtual Reality in Education & training, human factors in VR, Augmented Reality in Entertainment and Education.

**Dr. P. Dhanapal Durai Dominic** obtained his M.Sc degree in operations research in 1985, MBA from Regional Engineering College, Tiruchirappalli, India during 1991, Post Graduate Diploma in Operations Research in 2000 and completed his Ph.D during 2004 in the area of job shop scheduling at Alagappa University, Karaikudi, India. Presently he is working as a Senior Lecturer, in the Department of Computer and Information Science, Universiti Teknologi PETRONAS, Malaysia. His fields of interest are Operations Management, KM, and Decisions Support Systems. He has published technical papers in International, National journals and conferences.